\documentclass[lettersize,journal]{IEEEtran}
\usepackage{amsmath,amsfonts}
\usepackage{algorithmic}
\usepackage{algorithm}
\usepackage{array}
\usepackage{url}
\usepackage[caption=false,font=normalsize,labelfont=sf,textfont=sf]{subfig}
\usepackage{textcomp}
\usepackage{stfloats}
\usepackage{url}
\usepackage{booktabs}
\usepackage{verbatim}
\usepackage{graphicx}
\usepackage{float}
\usepackage{cite}
\usepackage{multirow}
\usepackage{pifont}
\begin{document}

\title{STCGAT: A Spatio-temporal Causal Graph Attention Network for traffic flow prediction in Intelligent Transportation Systems}

\author{Wei Zhao, Shiqi Zhang, Bing Zhou, Bei Wang
	
	\thanks{W. Zhao is with the School of Artificial Intelligence and Computer Science, School of Cyber Science and Engineering and Cooperative Innovation Center of Internet Healthcare, Zhengzhou University, Zhengzhou, 450001, China.}
	\thanks{S. Zhang and B. Wang are with the School of Cyber Science and Engineering, Zhengzhou University, Zhengzhou, 450001, China(e-mail:zhangshiqi$ @ $gs.zzu.edu.cn).}
	\thanks{B. Zhou is with the School of Artificial Intelligence and Computer Science and Cooperative Innovation Center of Internet Healthcare, Zhengzhou University, Zhengzhou, 450001, China.}}

\markboth{Journal of \LaTeX\ Class Files,~Vol.~14, No.~8, August~2021}%
{Shell \MakeLowercase{\textit{et al.}}: A Sample Article Using IEEEtran.cls for IEEE Journals}


\maketitle

\begin{abstract}
Air pollution and carbon emissions caused by modern transportation are closely related to global climate change. With the help of next-generation information technology such as  Internet of Things (IoT) and Artificial Intelligence (AI), accurate traffic flow prediction can effectively solve problems such as traffic congestion and mitigate environmental pollution and climate change. It further promotes the development of Intelligent Transportation Systems (ITS) and smart cities. However, the strong spatial and temporal correlation of traffic data makes the task of accurate traffic forecasting a significant challenge. Existing methods are usually based on graph neural networks using predefined spatial adjacency graphs of traffic networks to model spatial dependencies, ignoring the dynamic correlation of relationships between road nodes. In addition, they usually use independent Spatio-temporal components to capture Spatio-temporal dependencies and do not effectively model global Spatio-temporal dependencies. This paper proposes a new Spatio-temporal Causal Graph Attention Network (STCGAT\footnote{The source code of STCGAT is available at \url{https://github.com/zsqZZU/STCGAT}.}) for traffic prediction to address the above challenges. In STCGAT, we use a node embedding approach that can adaptively generate spatial adjacency subgraphs at each time step without a priori geographic knowledge and fine-grained modeling of the topology of dynamically generated graphs for different time steps. Meanwhile, we propose an efficient causal temporal correlation component that contains node adaptive learning, graph convolution, and local and global causal temporal convolution modules to learn local and global Spatio-temporal dependencies jointly. Extensive experiments on four real, large traffic datasets show that our model consistently outperforms all baseline models.
\end{abstract}


\section{Introduction}
\IEEEPARstart{W}{ith} the rapid development of Industrial Internet of Things (IIoT) 4.0\cite{ref_2}, Healthcare 2.0, 5G, and even future 6G-based high-traffic communications\cite{ref_1}, as well as Artificial Intelligence (AI), the range of intelligent city sub-sectors are beginning to serve service providers and citizens fully. Among these services, Intelligent Transportation Systems (ITS)\cite{ref1} is the critical platform for intelligent transportation, which provides outstanding services. Such as, it can provide real-time and accurate road traffic status information, location navigation services, personalized travel route planning, and other services. As an essential part of the intelligent transportation system, traffic flow prediction can effectively mine potential Spatio-temporal patterns from traffic data, which not only helps to relieve traffic congestion and control traffic flow scheduling but also reduces people's travel time and cost and reduces environmental pollution to promote the development of smart cities\cite{ref2}.

The prediction model based on Spatio-temporal traffic data should consider the temporal dependencies of historical time series and the spatial dependencies between traffic highways. In the early days, traffic flow prediction methods were usually regarded as multivariate series analysis tasks in the time dimension, such as modeling traffic time series data using queuing theory\cite{ref3} model, traffic behavior theory\cite{ref4}, and machine learning approaches\cite{ref5}. However, these methodologies only evaluated the dependence on the temporal dimension and disregarded the spatial dimension dependence. Consequently, an increasing number of academics are focusing on Spatio-temporal prediction models based on Graphical Neural Network (GNN)\cite{ref6} that produce impressive outcomes. However, these models still have some limitations.

\begin{figure}[htb]
	\centering
	\includegraphics[width=3.5in]{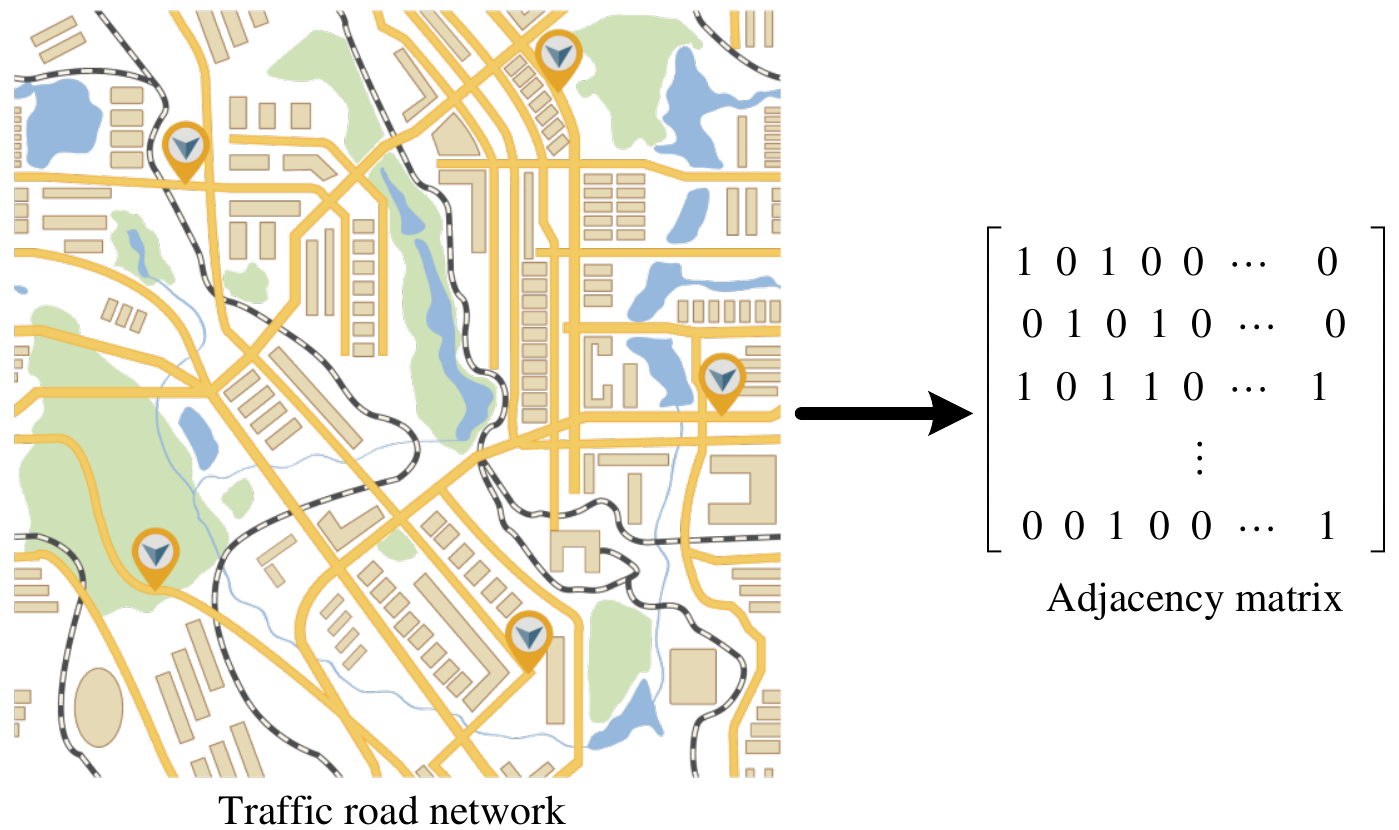}
	\caption{Adjacency Matrix of Traffic Road Network.}
	\label{fig1}
\end{figure}

The first limitation is that the dynamic correlation information between nodes on the graph is ignored. The GNN-based spatial dependency modeling approach can be thought of as edge transformation and aggregation through the information of nodes in the traffic network\cite{ref6}. As shown in Fig. \ref{fig1}, Most approaches\cite{ref7}\cite{ref8}\cite{ref9} employ a predetermined static adjacency matrix to characterize the spatial interactions between traffic road nodes\cite{ref10}. However, the relationships between road nodes in a traffic network are dynamic and interact, depending on various complex factors on the road, such as traffic flow, number of lanes, and population density. Therefore, the spatial relationship of roads cannot be modeled simply by the static spatial connections between them.

Second, traffic information on the transportation network has a high degree of nonlinear correlation and uncertainty. Examples include regular road maintenance, sudden accidents, etc. However, These models usually use Recurrent Neural Network (RNN)\cite{ref11} (eg. Long-short Term Memory (LSTM)\cite{ref12} or Gated Recurrent Unit (GRU)\cite{ref13}) based models to capture the temporal dependence, the signals must be traversed along the long recurrent path of the network when dealing with long-range sequences, making it challenging to effectively model the global time dependence of long-time sequence data. In addition, the sequential execution process of RNNs prevents them from capturing causal correlation information about traffic events\cite{ref14}. To address the above issues, some Convolutional Neural Network (CNN)-based research approaches stack convolutional layers into multiple layers to model global temporal dependencies\cite{ref15}\cite{ref16}. However, the local time-dependent information may be lost as the expansion rate increases\cite{ref17}. Meanwhile, as the neural network's depth deepens, it makes optimizing the model more challenging, which inevitably leads to the problem of network degradation\cite{ref18}.

To address the aforementioned challenge, we offer a new Spatio-temporal prediction framework based on Graph Attention Network (GAT)\cite{ref19} called Spatio-temporal Causal Graph Attention Network (STCGAT). We propose a data-driven graph structure learning method that can autonomously learn the relationship information between road nodes and model the spatial correlation of traffic networks without relying on the traffic network graph structure information. In addition, we propose a bidirectional Spatio-temporal component to simultaneously capture local and global spatial-temporal dependency information. Moreover, a residual module\cite{ref18} is introduced in the component to use a deep network to capture more abstract Spatio-temporal dependence information. The following are the principal contributions of this work:

\begin{enumerate}
	\item{We construct a data-driven Node Adaptive Learning Graph Attention Network (NAL-GAT) that dynamically captures spatial dependencies by generating a weighted adjacency matrix of the graph based on the road traffic state information at different moments.}
	\item{We propose a new spatiotemporal component that replaces the GRU gating unit with NAL-GAT and is further constructed as a recursive bidirectional network to capture the local causal spatiotemporal dependence.}
	\item{We parallel process the output sequence data by stacked Temporal Convolutional Network to capture global and long-range temporal dependencies.}
\end{enumerate}

The structure of this document is as follows: 
Section \uppercase\expandafter{\romannumeral2} briefly covers some relevant works in the field of spatial-temporal prediction. 
Section \uppercase\expandafter{\romannumeral3} explains the structure and methodology of STCGAT. 
Section \uppercase\expandafter{\romannumeral4} evaluates the performance of STCGAT by comparing numerous experimental outcomes. 
Section \uppercase\expandafter{\romannumeral5} finally closes our job.

\begin{figure*}[h]%
	\centering
	\includegraphics[width=0.9\textwidth]{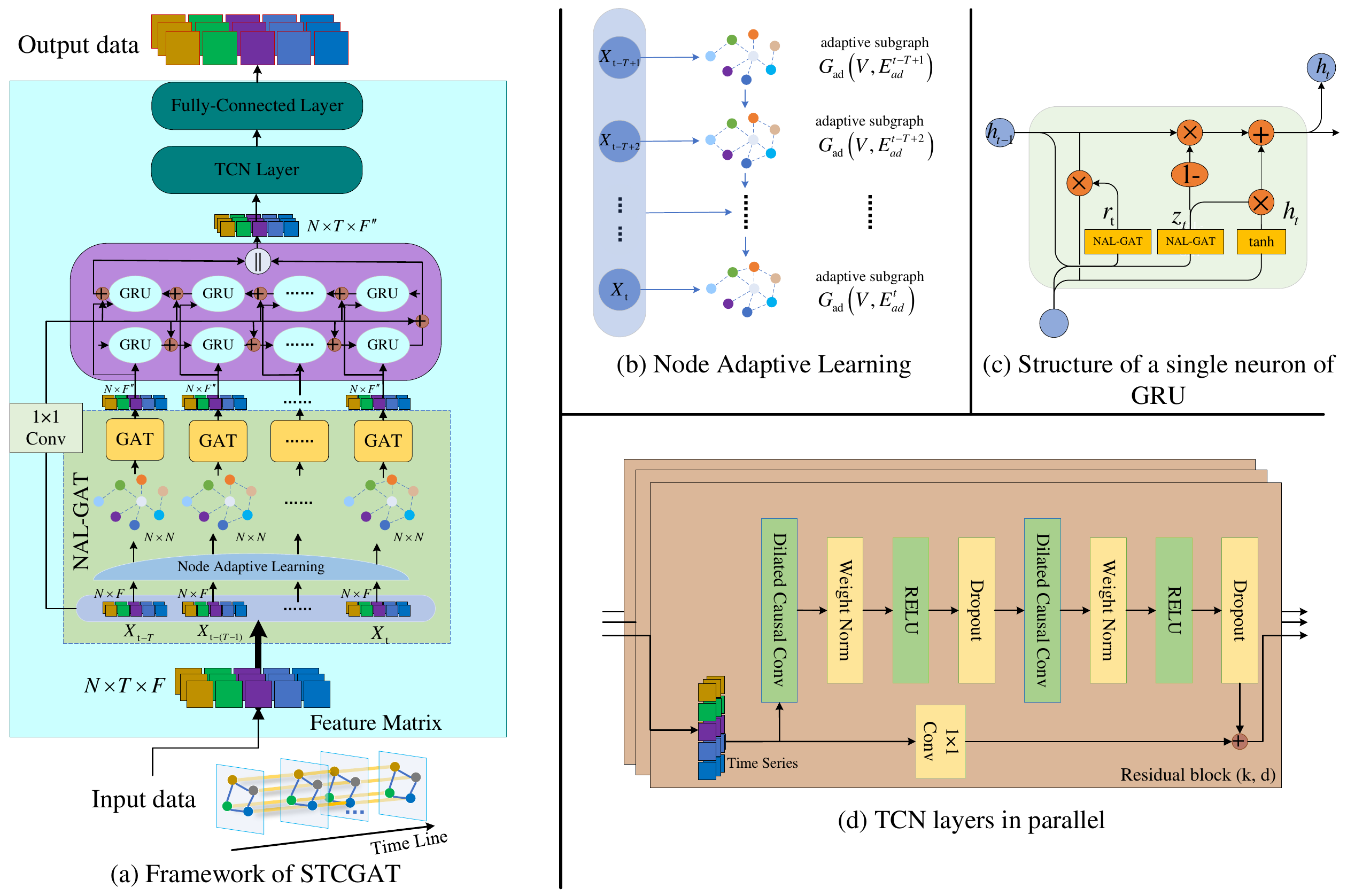}
	\caption{The STCGAT framework in its entirety. (a) This is the general structure diagram of STCGAT, which uses past traffic data and a sophisticated temporal component to anticipate the traffic situation for a future period. (b) This is the node-adaptive learning mechanism, which constructs the relational subgraph $G_{ad}$ of the traffic road network adaptively using a learnable road node embedding matrix that learns the edge relationship set $E_{ad}$ of nodes for traffic status information on traffic roads at different times. Moreover, it is coupled with GAT to capture the spatial dependence adaptively. (c) The gating unit of GRU is replaced with NAL-GAT to model both spatial and temporal dependence. (d) This is the residual module of the temporal convolution layer, in which TCN processes local time series data in parallel to capture global Spatio-temporal dependency.  Where $k$ is the size of the filter and $d$ represents the expansion factor.}
	\label{fig2}
\end{figure*}

\section{Related Works}
\subsection{Graph Convolutional Network}
Graph Convolutional Networks (GCN)\cite{ref20} has received much interest in recent years. They are now frequently utilized for various tasks involving graph-structured data, including graph classification, node classification, and link prediction. The clustering method divides GCN into two basic categories: Spectral-based and Spatial-based\cite{ref6}. Spectral-based techniques implement filters to define the graph convolution from the perspective of graph signal processing and extend the graph convolution to the spectral domain by locating the corresponding Fourier bases. The primary approaches are Chebyshev Spectral CNN (ChebNet)\cite{ref21} and Adaptive Graph Convolution Network (AGCN)\cite{ref22}. The Spatial-based approach represents graph convolution as aggregating feature information from the neighborhood and defining the graph convolution by the spatial relationship of the nodes. The most prominent methods are GAT and Gated Attention Network (GAAN)\cite{ref23}.

\subsection{Time-dependent modeling}
Earlier traffic forecasting projects were often multivariate time series analysis problems utilizing time series modeling techniques such as History Average (HA) Model\cite{ref24}, Vector Autoregression (VAR)\cite{ref25} and Support Vector Regressor (SVR)\cite{ref26}. However, these algorithms rely too heavily on the assumption of ideal smoothness, which is incompatible with the nonlinear correlation of traffic data. With its well-established data modeling and autonomous learning capabilities, deep learning has steadily taken over time series forecasting tasks in recent years. Most studies have used RNN-based models to capture temporal dependencies, but the complex long-loop structure is time-consuming. It may also be accompanied by gradient disappearance and explosion phenomena as the long-loop links grow. To address this problem, some work uses Temporal Convolutional Network (TCN)\cite{ref27}\cite{ref28} to process time series in parallel, learning more helpful information through a larger field of perception. Recently, several researchers have concentrated on the robust modeling capabilities of Transform\cite{ref29} for time series data and have presented many variant models\cite{ref30}\cite{ref31}\cite{ref32} based on Transform with excellent performance on long time series prediction tasks.

\subsection{Spatio-temporal dependence modeling}
To model traffic data's spatial and temporal dependencies, some studies\cite{ref14}\cite{ref15}\cite{ref16} modeled traffic networks as regular two-dimensional grids, then process the two-dimensional data using CNN to capture the spatial dependence, and finally further model the temporal dependence of traffic time series using RNN or CNN. However, CNN's are not always applicable in traffic road networks of non-Euclidean nature. To solve the problem, more and more researchers are turning to GCN-based models for Spatio-temporal data prediction. For example, DCRNN\cite{ref7} models spatial dependencies by wandering in both directions on the traffic road topology graph and then uses GRU to capture temporal dependencies. ASTGCN\cite{ref8} models the spatial-temporal dependence of traffic data with spatial attention and temporal attention, respectively. STSGCN\cite{ref33} captures the local Spatio-temporal correlation of traffic data by combining Spatio-temporal blocks through a local Spatio-temporal synchronized graph convolution module. STFGNN\cite{ref17} learns both local and global Spatio-temporal dependencies by processing data-driven spatial and temporal graphs at different moments in parallel. Following STFGNN, STGODE\cite{ref34} draws on the Dynamic Time Warping (DTW)\cite{ref35} used by STFGNN to generate semantic adjacency matrices for traffic road topology maps to capture deeper Spatio-temporal dependencies. In recent works, Z-GCNETs\cite{ref36} introduce the concept of zigzag persistence in the traffic network diagram structure and integrate it into GCNs to enhance the stability of the model. STG-NCDE\cite{ref37} designs two neural control differential equations dealing with temporal and spatial dependencies, respectively, and integrates both to capture Spatio-temporal dependencies simultaneously.

\section{Methods}
\subsection{Problem formulation}

Traffic flow forecasting uses historical traffic information on the road to predict traffic condition information in the future period. In our model, traffic speed is chosen as the traffic condition information of the road.

\textbf{Definition 1}: We use the undirected graph $G=(V,E)$ to represent the topological information of the traffic roads. Where $V=\{v_1,v_2,\cdots,v_N\}$ is the set of all nodes of graph $G$, $N$ denotes the number of all road nodes, and $E$ denotes the set of connected edges of all nodes of graph $G$.

\textbf{Definition 2}: We represent the historical traffic speed information of time length $T$ by a feature matrix $X=\{ X_1,X_2, \cdots, X_N\}\in R^{N \times T \times F}$, where $F$ denotes the feature dimension, $X_t=\{ \vec{x_{t:1}},\vec{x_{t:2}}, \cdots, \vec{x_{t:N}}\} \in R^{N \times F}$  denotes the set of traffic information of all road nodes at any $t$ time, and $\vec{x_{t:i}} \in \mathbb{R}^F$ then denotes all the feature vectors of node $v_i$.

Traffic prediction has a substantial spatial and temporal correlation, considering the spatial correlation and temporal dependence of individual road nodes in the road network. Therefore, we aim to use the road network topology $G$ and the traffic information feature matrix $X$ to predict the traffic speed information $Y^\prime=[X_{t+1}^\prime,X_{t+2}^\prime,\cdots,X_{t+T}^\prime]$ in the next $T$ moments by learning a function  $f (\cdot)$.

\begin{equation}
	\label{deqn_ex1}
	Y^\prime=f (G;(X_{t-T},X_{t-(T-1)},\cdots,X_{t}))
\end{equation}
\subsection{Spatial Dependency Modeling}
The traffic condition data of distinct road nodes are strongly and dynamically connected in the spatial dimension. However, traditional graph neural networks construct predefined adjacency matrices to perform graph convolution operations based on information such as connectivity or distance between graph nodes. Although the predefined adjacency matrix intuitively represents the position relationship between nodes, it cannot reflect the dynamic spatial dependence between road nodes at different moments dynamically. Therefore, as shown in (b) in Fig. \ref{fig2}, we use a node-adaptive learning mechanism to learn the dynamic correlation information among road nodes at different moments in a fine-grained manner to generate the traffic subgraph $G_{ad}$ for the corresponding moment. As shown in Eq. \ref{deqn_ex2}, this mechanism generates the adjacency matrix $\tilde{A^t} \in R^{N \times N}$ corresponding to that moment at any moment $t$.

\begin{equation}
	\label{deqn_ex2}
	\tilde{A}^t=softmax(ReLU(E_{At} \cdot E_{At}^T))
\end{equation}
Where $E_{At} \in R^{N \times d}$ is the embedding dictionary encoding each traffic road node, $d$ is the embedding dimension, and $E_{At}^T$ is the transpose matrix of $E_{At}$. $softmax$ is the normalization function, and $ReLU$ is the nonlinear activation function.

Meanwhile, to capture spatial dependencies among nodes in the spatial dimension adaptively and dynamically. We propose a NAL-GAT to extract the spatial features of traffic roads by integrating the node-adaptive learning mechanism with GAT. Specifically, at any moment $t$, GAT computes the attention coefficients of its neighboring nodes vertex by vertex for the node correlation information generated by the node adaptive learning mechanism. Finally, it aggregates the spatial dependencies among the nodes on the graph at moment $t$. As shown in Eq. \ref{deqn_ex3}, the computation process of attention coefficient $e_{ij}^t$ between node $v_i$  and its arbitrary neighbor node $v_j$ is demonstrated.

\begin{equation}
	\label{deqn_ex3}
	e_{ij}^t=a(W\vec{x}_{t:i}, W\vec{x}_{t:j})
\end{equation}
Where $a$ is the computational function of the attention mechanism, and $W\in R^{F \times F^\prime}$ is the graph's weight matrix of all nodes. The attention coefficient $\alpha_{ij}^t$ of the graph's attention layer is then generated by normalizing the attention coefficients of node $v_i$'s neighbors.

\begin{equation}
	\label{deqn_ex4}
	\begin{split}
	\alpha_{ij}^t&=softmax_j(e_{ij}^t) \\
	&=\frac{exp(e_{ij}^t)}{\sum_{k \in \tilde{A}_i^t} exp(e_{ik}^t)}
	\end{split}
\end{equation}
Where $\tilde{A}_i^t$ denotes all the neighbor nodes of $v_i$.

In addition, we note that all nodes in the GAT share the same parameter space $W\in R^{F \times F^\prime}$, which may lead to too large graph $W$ with more nodes making the model difficult to optimize. To solve this problem, we construct a shared weight pool $W_p \in R^{d\times F \times F^\prime}$, which can get the weight matrix $W^\prime=E_{At}\cdot W_p \in R^{F\times F^\prime}$ of each node according to the node's embedding dictionary $E_{At}$. Then the complete attention coefficient calculation process is shown in Eq. \ref{deqn_ex5}.

\begin{equation}
	\label{deqn_ex5}
	\begin{split}
		e_{ij}^t&=LeakReLu(\vec{a}^T[E_{At}\cdot W_p\vec{x}_{t:i} \parallel E_{At}\cdot W_p\vec{x}_{t:i}]) \\ 
		e_{ik}^t&=LeakReLu(\vec{a}^T[E_{At}\cdot W_p\vec{x}_{t:i} \parallel E_{At}\cdot W_p\vec{x}_{t:k}]) \\
		\alpha_{ij}^t&=\frac{exp(e_{ij}^t)}{\sum_{k \in \tilde{A}_i^t} exp(e_{ik}^t)} \\
	\end{split}
\end{equation}
Where $\vec{a} \in\mathbb{R} ^{2F^\prime}$ is the weight matrix, $\parallel$ denotes the connection operation, and $LeakReLu$ is the nonlinear activation function.

To capture deeper feature information, this paper further uses the multi-head attention mechanism to model spatial dependence. As shown in Eq. \ref{deqn_ex6}, $Q$ sets of mutually independent attention mechanisms are invoked.

\begin{equation}
	\label{deqn_ex6}
	\vec{x}_{t:i}^\prime=\parallel_{q=1}^QLeakReLu(\sum_{k \in \tilde{A}_i^t}\alpha_{ik}^{t,q}(E_{At}^q\cdot W_p^q)\vec{x}_{t:k})
\end{equation}
Where $\alpha_{ik}^{t,q}$ is the weight coefficient computed by the attention mechanism of the qth group at time $t$, $E_{At}^q\cdot W_p^q$ is the weight matrix of the corresponding group, and $\vec{x_i}^\prime \in \mathbb{R}^{QF^\prime}$ is the new feature representation obtained by passing node $v_i$ through the attention layer of the multi-headed graph.

When using multi-headed attention, the network's last layer should not output too many features, so we use a separate self-attentive mechanism to limit the node's output feature length. Specifically, we use a new weight pool $W_p^\prime \in R^{d\times QF^\prime \times F^{\prime\prime}}$ to map the output dimension of nodes from $\mathbb{R}^{QF^\prime}$ to $\mathbb{R}^{F^{\prime\prime}}$ to obtain the final output result $\vec{x}_{t:i}^{\prime\prime} \in \mathbb{R}^{F^{\prime\prime}}$.

\begin{equation}
	\label{deqn_ex7}
	\vec{x}_{t:i}^{\prime\prime} = LeakReLu(\sum_{k \in \tilde{A}_i^t}\alpha_{ik}^{t\prime}(E_{At}^\prime\cdot W_p^\prime)\vec{x}_{t:k}^{\prime})
\end{equation}

When all nodes in the graph have completed the above graph attention layer operation, we can obtain the output features $X_t^{\prime\prime}=\{ \vec{x}_{t:1}^{\prime\prime},\vec{x}_{t:2}^{\prime\prime}, \cdots, \vec{x}_{t:N}^{\prime\prime}\} \in R^{N \times F^{\prime\prime}}$. For the convenience of presentation, we express this process in Eq. \ref{deqn_ex8}.

\begin{equation}
	\label{deqn_ex8}
	X_t^{\prime\prime} =\sigma(\tilde{A}^tX_t(E_{At}\cdot W_p))
\end{equation}
where $\sigma(\cdot) $ is the computation function for the graph attention layer.

\subsection{Local Causal Spatial-temporal Dependency Modeling}
There is a correlation between traffic conditions in the time dimension at various times. As shown in (C) in Fig. \ref{fig2}, we replace the gating unit of GRU with NAL-GAT to further capture the Spatio-temporal dependence of the traffic time series data. Specifically, the spatially dependent time series data $X_t^{\prime\prime}$ at any moment $t$ is used as the input data of the GRU.

\begin{equation} \label{deqn_ex9}
	\begin{split}
		z_t&=\sigma(\tilde{A}^t[X_t, \overrightarrow{h}_{t-1}](E_{At}^z\cdot W_p^z)) \\ 
		r_t&=\sigma(\tilde{A}^t[X_t, \overrightarrow{h}_{t-1}](E_{At}^r\cdot W_p^r)) \\
		\widetilde{h_t}&=tanh(\tilde{A}^t[X_t, r_t \odot \overrightarrow{h}_{t-1}](E_{At}^{\widetilde{h}t}\cdot W_p^{\widetilde{h}t}) \\ 
		\overrightarrow{h_t} &=z_t\odot h_{t-1} + (1-z_t)\odot \widetilde{h_t}  
	\end{split}
\end{equation}
Where $\overrightarrow{h}_{t-1}$ is the output at the previous moment, $[\cdot]$ denotes the concat operation in the feature dimension, $\widetilde{h_t}$ is the candidate hidden layer state, $\odot$ denotes the multiplication by elements, and $\overrightarrow{h_t} \in R^{N \times F^{\prime\prime}}$ is the output at the current moment.

It is important to note that as the input time length rises, so does the network depth of the model. However, the deep network may lead to issues such as gradient disappearance and overfitting in the model. Therefore, we use the residual module to connect the network's layers to improve the model's capacity for long-term capture.

\begin{equation} \label{deqn_ex10}
	\begin{split}
		z_t&=\sigma(\tilde{A}^t[X_t, \overrightarrow{h}_{t-1}^\prime](E_{At}^z\cdot W_p^z)) \\
		r_t&=\sigma(\tilde{A}^t[X_t, \overrightarrow{h}_{t-1}^\prime](E_{At}^r\cdot W_p^r)) \\
		\widetilde{h_t}&=tanh(\tilde{A}^t[X_t, r_t \odot \overrightarrow{h}_{t-1}^\prime](E_{At}^{\widetilde{h}t}\cdot W_p^{\widetilde{h}t})\\
		\overrightarrow{h_t} &=z_t\odot h_{t-1}^\prime + (1-z_t)\odot \widetilde{h_t} \\
		\overrightarrow{h}_t^\prime &=\varepsilon(\omega_1 \otimes X_t + \omega_2 \otimes \overrightarrow{h_t})
	\end{split}
\end{equation}
Where $\omega_1$ and $\omega_2$ are both one-dimensional convolution kernels, $\varepsilon$ is the nonlinear activation function, $\otimes$ denotes the convolution operation, and $\overrightarrow{h}_t^\prime\in R^{N \times F^{\prime\prime}}$ is the output of residual concatenation. Until the completion of the above operations at the Tth time step, we can obtain the sequence data containing the Spatio-temporal dependence $\overrightarrow{H}^\prime=(\overrightarrow{h}_{t-T}^\prime,\overrightarrow{h}_{t-(T-1)}^\prime ,\cdots,\overrightarrow{h}_{t}^\prime)$. 

In addition, traffic data are not always sequentially correlated, and there are complex causal correlations between traffic events. Therefore, we use bidirectional GRU to capture the local causal, temporal relationships. The reverse operation is similar to the above operation, and the output results are finally stitched to obtain the output $H\in R^{N \times T \times 2F^{\prime\prime}}$.

\begin{equation}
	\label{deqn_ex11}
	H = \overrightarrow{H}^\prime \parallel \overleftarrow{H^\prime}
\end{equation}
\subsection{Global Spatial-temporal Dependency Modeling}
From the above procedure, it is evident that GRU is processed by progressively unfolding along the timeline, which causes the output at the present time to depend on the state at the previous time and so lacks the capacity to capture long-term temporal dependence. We deploy a parallel temporal convolutional network along the time axis to improve the performance of extracting long-term temporal dependencies. The output of Equation 11's data deformation $H^\prime\in R^{N \times (T * 2F^{\prime\prime}})$ is utilized as the TCN's input data. In the time series convolution process, the time series data $H_{i:}^\prime \in \mathbb{R}^{(T*F^{\prime\prime})}$ of any node $v_i$ and a filter $f:\{0,\cdots, l-1\}\ \ \overrightarrow{}\ \ \mathbb{R} $ are first extended for the elements $s$.

\begin{equation}
	\label{deqn_ex12}
	F(s)=(H_{i:}^\prime *_df)(s)=\sum_{i=0}^{l-1}f(i)\cdot H_{i:}^\prime (_{s-d \cdot i})
\end{equation}
Where $d$ is the dilation factor, $l$ is the filter size, and $s-d \cdot i$ denotes the direction of the timeline past. When $d=0$, the dilation convolution becomes a regular convolution. In addition, as shown in (d) in Fig. \ref{fig2}, the TCN needs to perform a series of transformations such as Weight Norm and Dropout and use the residual join to obtain the output $o \in R^{N \times (T * 2F^{\prime\prime})}$ as in Eq. \ref{deqn_ex13}.

\begin{equation}
	\label{deqn_ex13}
	o = Activation(x + F(x))
\end{equation}

\subsection{The prediction layer}
Finally, we perform a dimension-specific linear transformation of the output sequence by a two-layer fully connected neural network.
\begin{equation}
	\label{deqn_ex14}
	Y^\prime =W_2\cdot\varphi(W_1 \cdot o +b_1) +b_2
\end{equation}
Where $W_1\in R^{F^{\prime\prime\prime} \times (T * 2F^{\prime\prime})}$ and $W_2\in R^{(T\times F) \times F^{\prime\prime\prime}}$ are the weight matrices, $b_1$ and $b_2$ are the corresponding bias terms, and $Y^\prime \in R^{N \times T \times F}$ is the final prediction result.

During model training, we optimize the model using the $L1$ loss function and the Adam optimizer to make the error between the predicted $Y^\prime=[X_{t+1}^\prime,X_{t+2}^\prime,\cdots,X_{t+T}^\prime]$ and labeled values $Y=[X_{t+1},X_{t+2},\cdots,X_{t+T}]$ as small as possible.
\begin{equation}
	\label{deqn_ex15}
	loss =\frac{1}{T}\sum_{i=1}^{T}\lvert Y_{t+i} - Y_{t+i}^\prime \rvert
\end{equation}

\section{Experiment}
\subsection{Datasets}
We have done extensive experiments on the following four real public transportation datasets, PeMS03, PeMS04, PeMS07, and PeMS08\cite{ref38}, to illustrate the effectiveness of the proposed Spatio-temporal prediction framework. These datasets were obtained on Caltrans' Performance Measurement System (PeMS). The system has over 39,000 traffic detectors deployed on California freeways to collect real-time traffic flow data and geographic information on the roadways. The collected data are aggregated every five minutes. Table \ref{tab1} summarizes the critical statistics for the four datasets.
\begin{table}[h]
	\begin{center}
		\renewcommand{\arraystretch}{1.4}
		\caption{Datasets information statistics.}
		\label{tab1}
		\begin{tabular}{| c | c | c |c | c |}
			\hline
			Datasets & Sensors  & Edges & Unit & Time Steps\\
			\hline
			PeMS03    & 358&547&5 min&26208\\
			
			\hline
			PeMS04    & 307&340&5 min&16992\\
			\hline
			PeMS07    & 883&866&5 min&28224\\
			\hline
			PeMS08    & 170&295&5 min&17856\\
			\hline
		\end{tabular}
	\end{center}
\end{table}

\subsection{Baseline Methods}
STCGAT was compared to some of the most advanced baseline models. The following is a summary of these baselines.
\begin{itemize}
	\item
	HA\cite{ref24}: The model uses the average of historical traffic data as the forecast value.
	\item
	VAR\cite{ref25}: The model captures the dependencies between temporal data. 
	\item
	FC-LSTM\cite{ref12}: The model is based on the traditional RNN model to model the time dependence of historical traffic data.
	\item
	TCN\cite{ref27}: The model allows for processing long-time series data at a fraction of the cost.
	\item
	DCRNN\cite{ref7}: The model is based on GCN to capture spatial dependencies and uses an encoder-decoder architecture to capture temporal correlations.
	\item
	ASTGCN\cite{ref8}: The model uses spatial and temporal attention mechanisms to capture underlying Spatio-temporal patterns.
	\item
	STSGCN\cite{ref33}: The model captures local Spatio-temporal dependencies using spatial graph convolution and one-dimensional temporal convolution, respectively.
	\item
	AGCRN\cite{ref9}: The model proposes an adaptive graph convolution module that can autonomously learn the adjacency matrix of a traffic road network.
	\item
	STFGNN\cite{ref17}: The model effectively fuses multiple spatial and temporal graphs to learn the Spatio-temporal relationships hidden by traffic data.
	\item
	STGODE\cite{ref34}: The model captures deeper Spatio-temporal dependencies by expanding the perceptual field of the GCN.
	\item
	Z-GCNETs\cite{ref36}: The model is designed with a time-aware GCN to capture the complex Spatio-temporal dependencies in traffic data.
	\item
	STG-NCDE\cite{ref37}: The model is designed with two independent neural control differential equations for modeling spatial and temporal dependence.
\end{itemize}

\subsection{Experimental Settings and Evaluation Metrics}
We divide the used dataset into 60\% training set, 20\% validation set, and the remaining 20\% test set with Z-score standardization. The partitioned dataset is then processed through a sliding window of length $2T$, where the first $T$ time lengths of serial data are used as historical data, and the last $T$ time lengths of data are used as labeled values. Here, we set the size of $T$ to $12$. That is, one hour's historical traffic data is used to predict traffic data for the next hour.

Meanwhile, STCGAT is implemented using the PyTorch deep learning framework. In the hyperparameter settings of our model, the embedding feature dimension of the nodes is set to 10, the hidden layer size is 64, the number of multi-head attention mechanisms is set to 3, and the convolutional kernel size is set to 2. During the training process, the learning rate is set to 0.001, the batch size is set to 64, and the model is optimized using the Adam optimizer with a maximum number of iterations of $300$. The training environment of the model is shown in Table \ref{tab_model}.

\begin{table}[h]
	\begin{center}
		\begin{minipage}{184pt}
			\caption{Experimental environments.}\label{tab_model}%
			\begin{tabular}{@{}|l|l|@{}}
				\hline
				System    & Ubantu 18.04.6\\
				\hline
				CPU    & Intel Core i5-10500 @ 3.10GHz\\
				\hline
				GPU    & NVIDIA GeForce 2080Ti\\
				
				\hline
			\end{tabular}
		\end{minipage}
	\end{center}
\end{table}

We use the following three performance metrics to measure the model's predictive power.

\begin{itemize}
	\item
	Mean Absolute Error(MAE):
\end{itemize}
\begin{equation*}
	\text{MAE}=\frac{1}{L}{\overset{L}{\underset {i=1}{\sum}}}\lvert Y_i-Y^\prime_i\rvert
\end{equation*}
\begin{itemize}
	\item
	Root Mean Squared Error(RMSE):
\end{itemize}
\begin{equation*}
	\text{RMSE}=\sqrt{\frac{1}{L} \sum^L_{i=1}(Y_i-Y^\prime_i)}
\end{equation*}

\begin{itemize}
	\item
	Mean Absolute Percentage Error(MAPE):
\end{itemize}
\begin{equation*}
	\text{MAPE}=\frac{100 \%}{L} \overset{L}{\underset {i=1}{\sum}} \lvert\frac{Y_i-Y^\prime_i}{Y_i}\rvert
\end{equation*}
Where \(L\) denotes the total number of samples. The lower the values of the three indicators above, the greater the model's predictive accuracy. We conduct each experiment five times and then calculate the mean value as the test result.

\begin{table*}[ht]
	\begin{center}
		\renewcommand{\arraystretch}{1.6}
		\caption{STCGAT and baseline models' performance on PeMS03, PeMS04, PeMS07, and PeMS08 datasets were compared.}
		\label{tab2}
		\resizebox{\textwidth}{!}{
			\begin{tabular}{cccccccccccccc}
				\hline
				\multirow{2}{*}{Model} & Dataset & \multicolumn{3}{c}{PeMS03}  & \multicolumn{3}{c}{PeMS04}    & \multicolumn{3}{c}{PeMS07}  & \multicolumn{3}{c}{PeMS08}   \\ \cline{2-14}
				
				& Metrics & MAE      & RMSE     & MAPE     & MAE      & RMSE     & MAPE & MAE      & RMSE     & MAPE     & MAE      & RMSE     & MAPE    \\
				\hline
				\multicolumn{2}{c}{HA}           & 31.74    & 51.79    & 33.49\%  & 39.87    & 59.04    & 27.59\% & 45.32    & 65.74    & 23.92\% & 35.16    & 59.74    & 28.35\%  \\
				\hline
				\multicolumn{2}{c}{VAR}          & 23.75    & 37.97    & 24.53\%  & 24.61    & 38.61    & 17.54\% & 49.89   & 75.45    & 32.13\% & 19.21    & 29.84    & 13.13\% \\
				\hline
				\multicolumn{2}{c}{FC-LSTM}      & 20.96    & 36.01    & 20.76\%  & 25.01    & 41.42    & 16.18\%  & 33.26    & 59.92    & 14.32\% & 23.49    & 38.89    & 14.55\% \\
				\hline
				\multicolumn{2}{c}{TCN}          & 19.32    & 33.55    & 19.93\%  & 23.22    & 37.26    & 15.59\% & 32.27    & 42.23    & 14.26\% & 22.72    & 35.79    & 14.03\% \\
				\hline
				\multicolumn{2}{c}{DCRNN}        & 17.48    & 29.19    & 16.83\%  & 21.22    & 33.44    & 14.17\% & 24.69    & 37.88    & 10.80\% & 16.82    & 26.32    & 10.92\% \\
				\hline
				\multicolumn{2}{c}{ASTGCN}       & 17.65    & 29.63    & 16.94\%  & 22.03    & 34.99    & 14.59\% & 24.01    & 37.87    & 10.73\% & 18.36    & 28.31    & 11.25\% \\
				\hline
				\multicolumn{2}{c}{STSGCN}       & 17.48    & 29.21    & 16.78\%  & 21.19    & 33.65    & 13.90\% & 24.26    & 39.03    & 10.21\% & 17.13    & 26.80    & 10.96\% \\
				\hline
				\multicolumn{2}{c}{AGCRN}        & \underline{15.97}    & 28.11    & \underline{15.23\%}   & 19.83    & 32.26    & 12.97\% & \underline{21.13}    & 35.20    & \underline{8.96\%} & 15.95    & 25.22    & 10.09\% \\
				\hline
				\multicolumn{2}{c}{STFGNN}       & 16.77    & 28.34    & 16.30\%  & 20.18    & 32.41    & 13.94\%  & 22.07    & 35.80    & 9.21\% & 16.64    & 26.25    & 10.60\% \\
				\hline
				\multicolumn{2}{c}{STGODE}       & 16.32    & 27.23    & 16.25\%  & 20.95    & 32.66    & 14.95\% & 22.90    & 37.54    & 10.14\% & 16.81    & 25.97    & 10.62\% \\
				\hline
				\multicolumn{2}{c}{Z-GCNETs}     & 16.64    & 28.15    & 16.39\%  & \underline{19.50}    & 31.61    & \underline{12.78\%} & 21.77    & 35.17    & 9.25\%& 15.76    & 25.11    & \underline{10.01\%}  \\
				\hline
				\multicolumn{2}{c}{STG-NCDE}     & 15.58    & \underline{27.08}    & 15.32\%  & 19.81    & \underline{31.57}    & 14.04\% & 21.78    & \underline{34.75}    & 9.42\% & \underline{15.62}    & \underline{24.87}    & 10.14\% \\
				\hline
				\multicolumn{2}{c}{$\mathbf{STCGAT}$}       & $\mathbf{15.31}$    & $\mathbf{27.03}$    & $\mathbf{14.90\%}$  & $\mathbf{19.21}$    & $\mathbf{31.12}$    & $\mathbf{12.36\%}$ & $\mathbf{20.89 }$   & $\mathbf{34.00}$   & $\mathbf{8.79\%}$& $\mathbf{15.41 }$   & $\mathbf{24.61}$    & $\mathbf{9.78\%}$  \\
				\hline
				
		\end{tabular}}
	\end{center}
\end{table*}

\subsection{Experiment Results and Analysis}
\begin{figure}[!ht]
	\centering
	\includegraphics[width=3.5in]{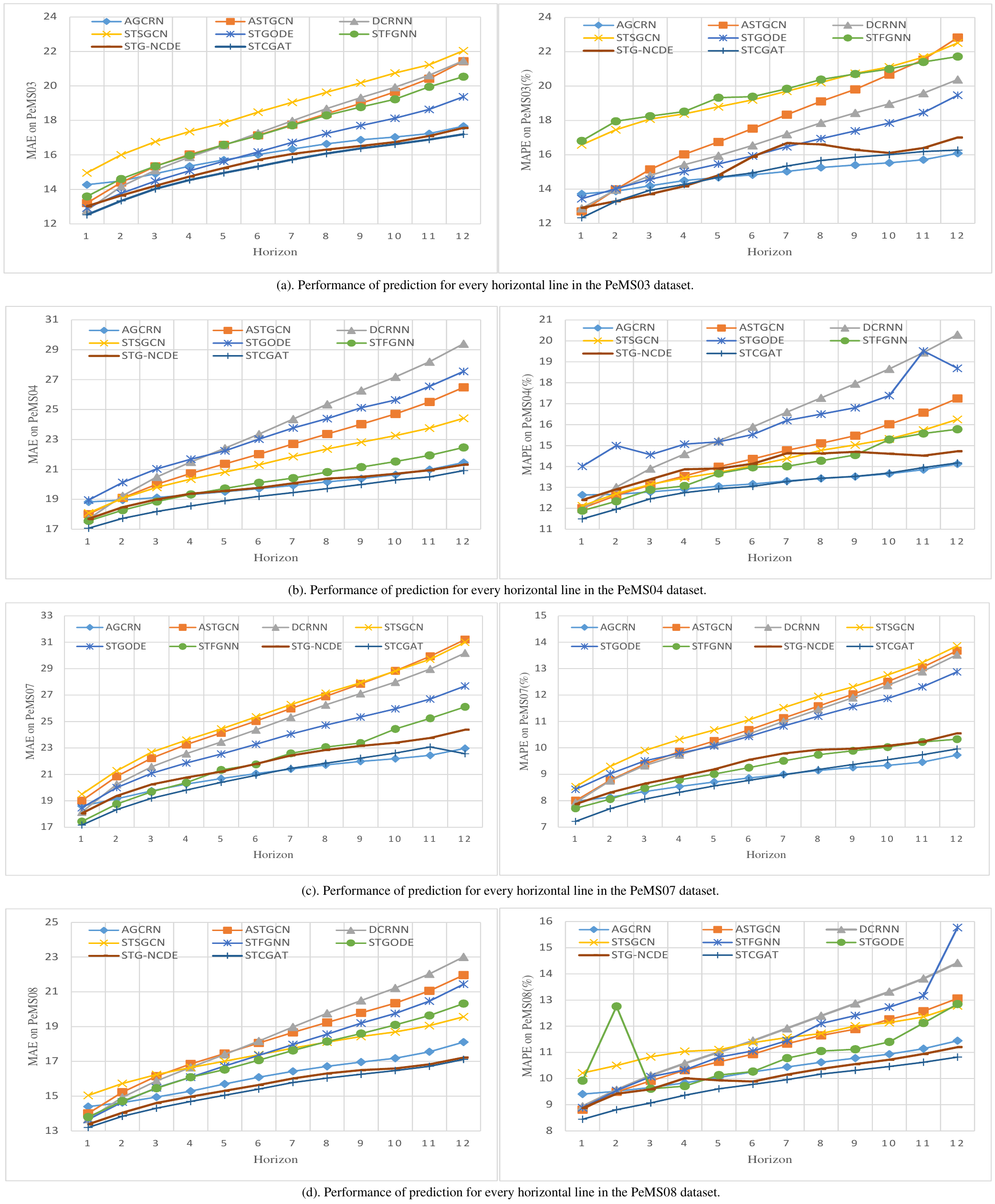}
	\caption{Metrics on PeMS03, PeMS04, PeMS07 and PeMS08 datasets.}
	\label{fig3}
\end{figure}
\begin{figure}[ht]
	\centering
	\includegraphics[width=3.5in]{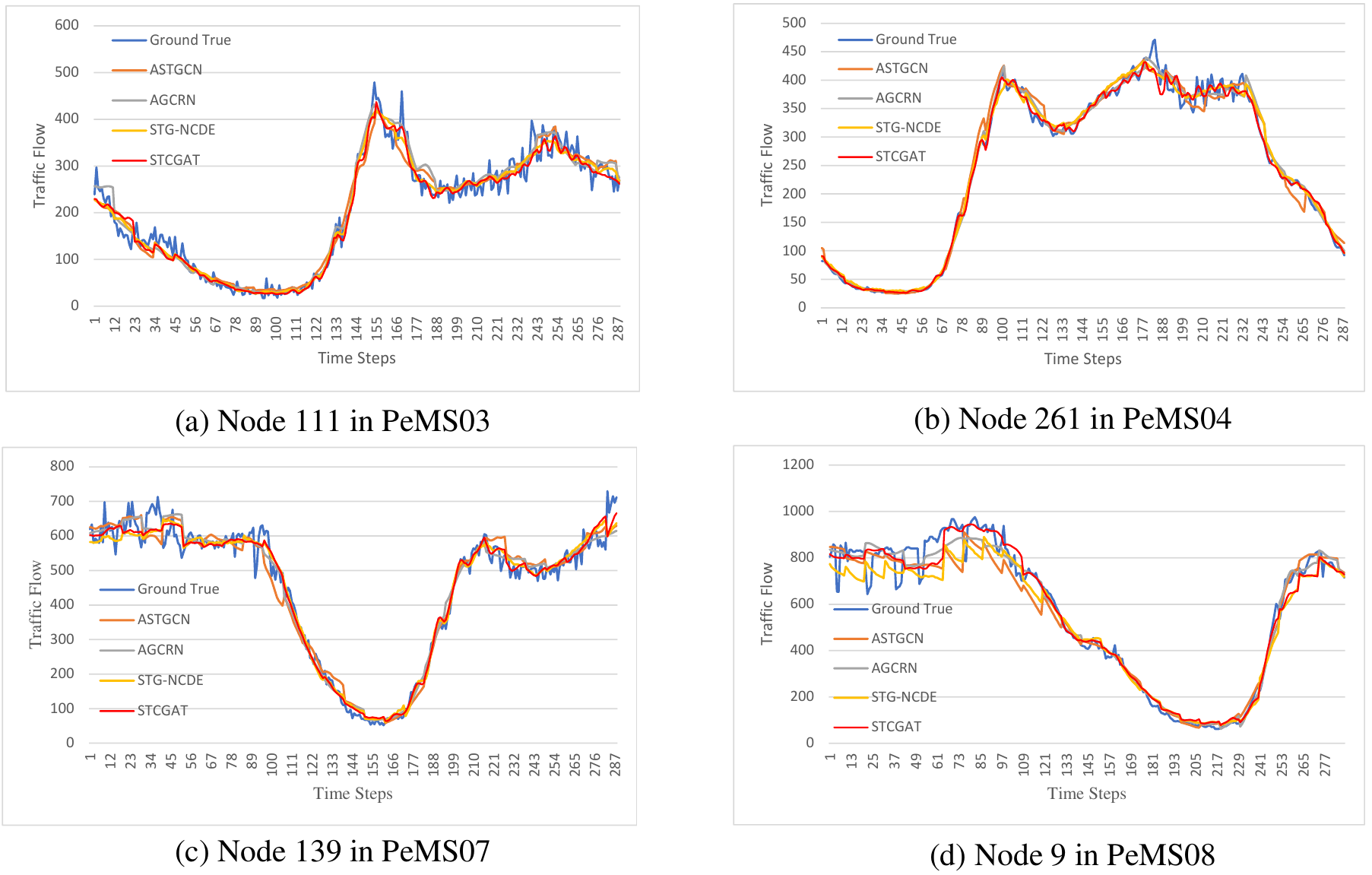}
	\caption{Traffic prediction visualization on PeMS03, PeMS04, PeMS07 and PeMS08 datasets.}
	\label{fig4}
\end{figure}

On PeMS03, PeMS04, PeMS07, and PeMS08, our model was compared with the twelve representative baseline approaches described previously. Table \ref{tab2} shows the prediction performance results of STCGAT and other baseline models within one hour (12 prediction steps). We can observe that 1) The GCN-based solution outperforms HA, VAR, and LSTM traditional machine learning methods, demonstrating the significance of explicitly modeling spatial correlation and the efficacy of GCN in traffic flow prediction tasks; 2) The performance metrics of our improved GAT-based method on each dataset are significantly better than other advanced baseline models, achieving significant results; 3) Our proposed method has stable short- and long-term Spatio-temporal prediction capabilities. As shown in Fig. \ref{fig3}, which demonstrates the comparison of the prediction performance of STCGAT with other Spatio-temporal prediction models on different Horizon, we can observe that the oscillations of the performance curve of STCGAT on each dataset are relatively small, indicating that our proposed method is insensitive to the prediction Horizon and the prediction performance is very stable; 4) As shown in Fig. \ref{fig4}, we use STCGAT and a representative baseline model to predict the traffic conditions at any road node for 288 consecutive time steps (24 hours). We can observe that the prediction results of STCGAT are more closely matched to the actual values, proving that STCGAT can more accurately capture the temporal and spatial correlations in the traffic flow sequence and achieve the best prediction results.

\begin{table*}[t]
	\begin{center}
		\renewcommand{\arraystretch}{1.4}
		\begin{minipage}{\textwidth}
			\caption{Results of STCGAT ablation experiments on PeMS04 and PeMS08 datasets.}\label{tab3}
			\begin{tabular*}{\textwidth}{@{\extracolsep{\fill}}lcccccc@{\extracolsep{\fill}}}
				\toprule%
				& \multicolumn{3}{@{}c@{}}{PeMS04} & \multicolumn{3}{@{}c@{}}{PeMS08} \\\cmidrule{2-4}\cmidrule{5-7}%
				Models & MAE & RMSE & MAPE & MAE & RMSE & MAPE \\
				\midrule
				w/o node embedding  & 22.03 & 34.65 & $15.29 \%$ & 17.68 & 27.65 & $11.04 \%$\\
				\hline
				w/o ResNet  & 19.72 & 32.45 & $13.57 \%$ & 16.01 & 25.37 & $10.31 \%$\\
				\hline
				w/o reverse GRU & 19.80 & 31.64 & $13.29 \%$ & 15.98 & 25.05 & $10.33 \%$\\
				\hline
				w/o TCN & 23.74 & 39.05 & $16.31 \%$ & 21.91 & 34.76 & $13.84 \%$\\
				\hline
				$\mathbf{STCGAT}$ & $\mathbf{19.21}$ & $\mathbf{31.12}$ & $\mathbf{12.36 \%}$ & $\mathbf{15.41}$ & $\mathbf{24.61}$ & $\mathbf{9.78 \%}$\\
				\hline
			\end{tabular*}
		\end{minipage}
	\end{center}
\end{table*}

\subsection{Ablation Study on Model Architecture}

We created four STCGAT-based model variants and contrasted STCGAT with these four variables on the PeMS04 and PeMS08 datasets to understand the influence of STCGAT's many modules. The differences between these four variants of the model are described below.
\begin{enumerate}
	\def\labelenumi{\arabic{enumi}.}
	\item
	w/o node embedding: The model removes the node embedding operation and uses only the traditional predefined adjacency matrix. 
	\item
	w/o ResNet: The model removes the residual connection structure in STCGAT.
	\item
	w/o reverse GRU: The model one-way GRU captures the Spatio-temporal correlation of the traffic network.
	\item
	w/o TCN: The model removes the temporal convolutional network.
\end{enumerate}

\begin{figure}[htb]
	\centering
	\includegraphics[width=3.5in]{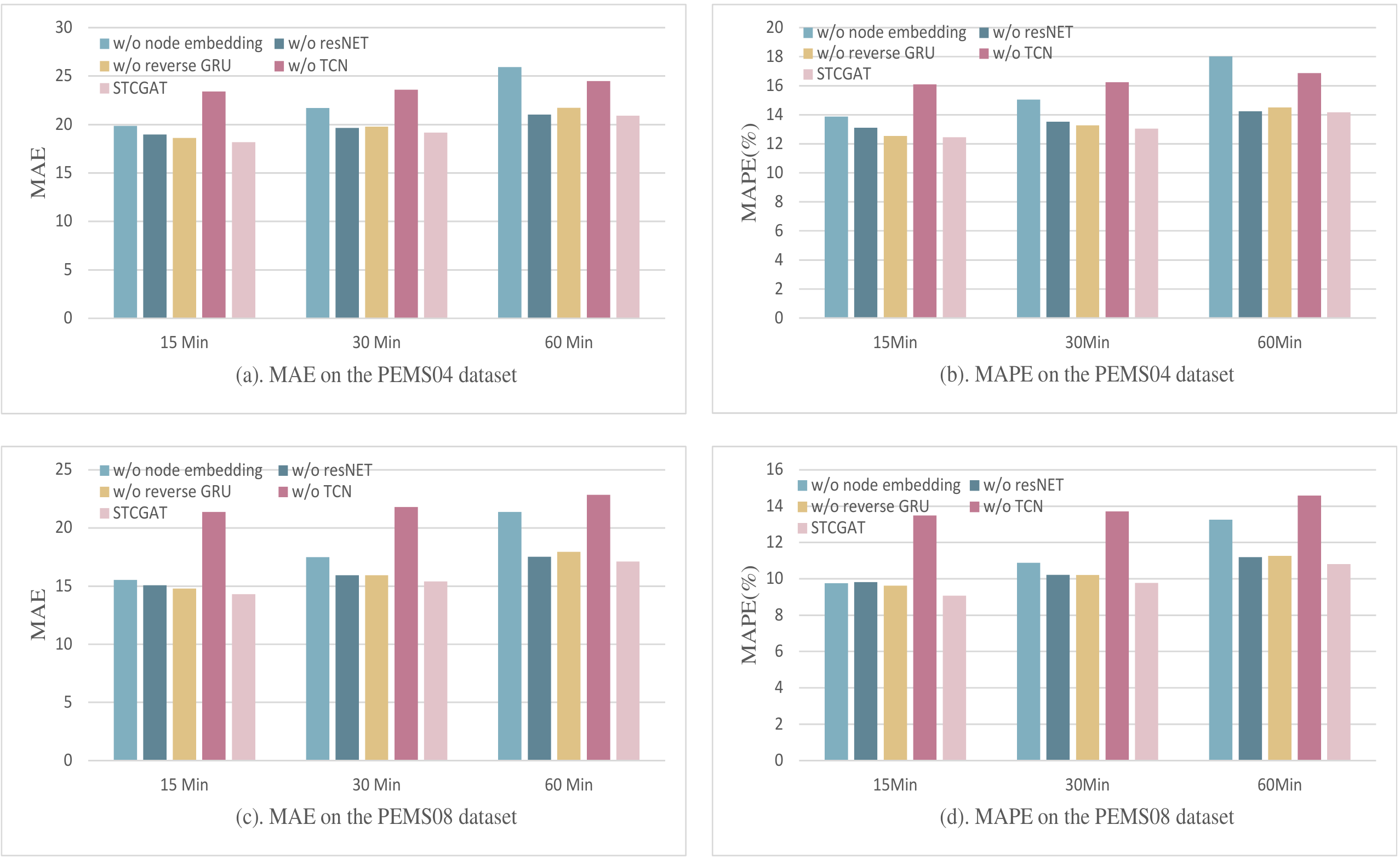}
	\caption{Short-term and long-term prediction performance of each variant model on the PeMS04 and PeMS08 datasets.}
	\label{fig5}
\end{figure}

Table \ref{tab3} shows that we conducted experiments with these four variants and STCGAT on PeMS04 and PeMS08. Combined with the histogram of evaluation metrics for each variant model at different time steps in Fig. \ref{fig5}, we can obtain the following observations: 1) Each evaluation metric with TCN is at its maximum value, and the performance metrics are similar in the short-term (15 min) and long term (60 min). This illustrates the necessity of using temporal convolutional networks to extract the global temporal correlation of traffic flow; 2) The performance metrics of w/o node embedding decrease significantly when the node self-learning module is removed. This proves that learning the dynamic spatial correlation among nodes from the state information of traffic flow at different moments better expresses how the traffic flow dynamics change in reality; 3) The performance metrics values of w/o ResNet compared to STCGAT are significantly larger, i.e., removing the residual module has a more significant impact on both datasets. This indicates that the application of the residual module helps STCGAT to mitigate the problem of overfitting or gradient disappearance caused by the superposition of network layers to a certain extent; 4) The performance metrics of with or without reverse GRU also increase on both the PeMS04 and PeMS08 datasets, indicating that effectively capturing the causality of traffic flow data helps to analyze the Spatio-temporal correlation of traffic flow more comprehensively; 5) Compared with the four variants, STCGAT has the best performance. On the one hand, it shows the importance of each module in STCGAT, and on the other hand, it shows that STCGAT can extract the Spatio-temporal correlations in traffic flow series more accurately.
\section{CONCLUSION}
This paper proposes a new Spatio-temporal causal prediction model STCGAT, which encodes all road nodes by node embedding. Then, we adaptively learn the relationship between road nodes according to the road traffic conditions at different moments, thus free from the constraint of the predefined adjacency matrix. It is further integrated into GAT to form NAL-GAT to model the spatial dependence of traffic road networks dynamically. Second, STCGAT reconstructs NAL-GAT into GRU to capture local Spatio-temporal dependencies. Meanwhile, STCGAT uses bi-directional GRU to capture the Spatio-temporal causality of traffic data in a fine-grained manner. In addition, STCGAT also reduces the network degradation problem caused by the deep network by introducing the residual module. Finally, STCGAT simultaneously captures traffic data's global Spatio-temporal dependence information by processing time series data in parallel with TCN. Through extensive comparative experiments on multiple datasets, it is demonstrated that STCGAT has excellent Spatio-temporal modeling capability for highly nonlinear traffic data and consistently achieves optimal prediction performance compared to advanced Spatio-temporal prediction models. In future work, we will investigate the proposed model in other spatiotemporal data mining problems, such as weather spatiotemporal data mining tasks.

\section*{Acknowledgement}
This work was supported in part by the National Key Technologies R\&D Program (2020YFB1712401, 2018YFB1701400), the 2020 Key Project of Public Benefit in Henan Province of China (201300210500), and the Nature Science Foundation of China (62006210).

\bibliographystyle{IEEEtran}

\bibliography{STCGAT.bib}

\vfill

\end{document}